\documentclass[letterpaper, 10 pt, conference]{ieeeconf}

\IEEEoverridecommandlockouts

\usepackage[utf8]{inputenc}
\usepackage[T1]{fontenc}
\usepackage{hyperref}
\usepackage{url}
\usepackage{booktabs}
\usepackage{amsfonts}
\usepackage{nicefrac}
\usepackage{microtype}
\usepackage{xcolor}
\usepackage{hhline}

\usepackage{multirow}
\usepackage{multicol}
\usepackage{booktabs}
\usepackage{amsmath}
\usepackage{amssymb}
\usepackage{graphicx}
\usepackage{nicefrac}
\usepackage{float}
\usepackage{wrapfig}
\usepackage{colortbl}

\usepackage{makecell}
\usepackage{wrapfig}
\usepackage{amssymb}
\usepackage{pifont}
\newcommand{\cmark}{\ding{51}}

\usepackage{xspace}
\usepackage{url}
\usepackage{capt-of}

\hypersetup{
	colorlinks = true,
	citebordercolor=.,
	citecolor=green,
}

\newcommand{\name}{\textit{GenRe}\xspace}
\newcommand{\namefull}{\textit{GenRe+}\xspace}

\title{\LARGE \bf
Diffusion-guided Generalizable Enhancer for Urban Scene Reconstruction
}

\author{Henry Che$^{1,3}$ \   Jingkang Wang$^{1,2}$ \ Yun Chen$^{1,2}$  \  Ze Yang$^{1,2}$  \   Sivabalan Manivasagam$^{1,2}$ \  Raquel Urtasun$^{1,2}$ \\
	\textsuperscript{1}Waabi  \quad
	\textsuperscript{2} University of Toronto \quad
	\textsuperscript{3}University of Illinois Urbana-Champaign \\
	\texttt{\small \{jwang, ychen, zyang, siva, urtasun\}@waabi.ai, hungdc2@illinois.edu
	}
}

\begin{document}

\twocolumn[{
	\renewcommand\twocolumn[1][]{#1}
	\maketitle
	\begin{center}
		\vspace{-6.0mm}
		\includegraphics[width=1.0\textwidth]{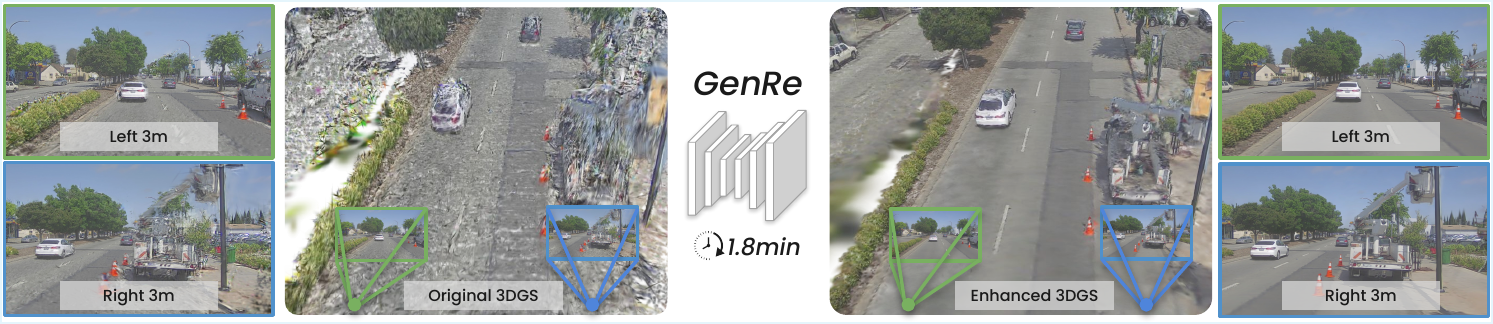}
	\end{center}
	\vspace{-4.0mm}
	\captionof{figure}{We introduce \name{}, a novel diffusion-guided generalizable enhancer for urban scene reconstruction.
		\name{} takes as input any pretrained 3D Gaussian representation and fixes the deficiencies within
		minutes,
		{producing robust, high-fidelity reconstructions that render reliably at novel viewpoints.}
		}
	\label{fig:teaser}
	\vspace{4.0mm}
}]

\thispagestyle{empty}
\pagestyle{empty}

\begin{abstract}
	Urban scene reconstruction from real-world observations has emerged as a powerful tool for self-driving 
	development and
	testing. While current neural rendering approaches achieve high-fidelity rendering along the recorded trajectories, their quality degrades significantly under large viewpoint shifts, limiting the applicability for closed-loop simulation. Recent works have shown promising results in using diffusion models to enhance quality at these challenging viewpoints and distill improvements back into 3D representations. However, they often require costly per-scene optimization, and the distilled representations remain fragile and fail to generalize beyond limited synthesized views. 
	To address these limitations, we propose \name{}, a novel diffusion-guided generalizable enhancer for urban scene reconstruction. \name{} takes as input any pretrained 3D Gaussian representation and fixes the deficiencies within a few minutes.
	By learning to distill generative priors across diverse scenes, \name{} produces robust and high-fidelity representation efficiently that generalizes reliably to challenging unseen viewpoints (\textit{e.g.}, 
	lane change). 
	Experiments show that \name{} outperforms existing methods in both quality and efficiency and benefits various downstream tasks, enabling robust and scalable sensor simulation for autonomous driving.
\end{abstract}
\section{Introduction}
\label{sec:intro}

Realistic simulation is essential to test safety-critical self-driving systems in a safe and scalable manner~\cite{wang2021advsim}.
Data-driven approaches, which construct digital twins
from real-world observations~\cite{lidarsim,wang2022cadsim}, have emerged as a key paradigm for sensor simulation. In contrast to artist-created,  game-engine–based virtual worlds~\cite{carla,airsim}, it provides scalability, realism, and diversity, forming a strong foundation for large-scale, closed-loop simulation for autonomy development.

Neural rendering approaches such as NeRF~\cite{nerf} achieve realistic reconstruction of urban driving scenes from camera and LiDAR data~\cite{yang2023unisim,tonderski2024neurad}, but are slow to render.
Recently, 3D Gaussian Splatting (3DGS)~\cite{3dgs} models scenes as large sets of explicit anisotropic Gaussians and renders them via rasterization, yielding faster rendering.
Subsequent works~\cite{yan2024street,chen2023periodic,chen2024omnire,splatad} extended this technique to dynamic urban driving scenes.
However, these differentiable-rendering pipelines often overfit to the training trajectories, leading to significant artifacts and quality degradation when extrapolating beyond original trajectory (\textit{e.g.}, meter-scale shifts).
In particular, 3DGS's over-parameterized primitives, in combination with the shape-radiance ambiguity when trained on single-trajectory ego views \cite{wu2025difix3d}, can exacerbate memorization of training views, producing floaters/holes and inconsistent surfaces under extrapolation (Fig.~\ref{fig:teaser} left).
Moreover, they cannot hallucinate plausible content in unobserved/occluded regions, resulting in holes and missing structures at extrapolated views.
These artifacts may reduce the fidelity of 
closed-loop sensor simulation, where the driving agent can deviate significantly from the recorded trajectory.

To address these limitations, recent work introduces physics-based and data-driven priors (\textit{e.g.},  additional regularization~\cite{vegs}, supervision from pre-trained vision models~\cite{peng2025desire}, shared decoders~\cite{splatad}, and generative models~\cite{vegs}) to stabilize the learned representation. Since these priors are not trained to handle reconstruction artifacts or representation-specific degradations, the gains are limited and often produce blurry results.
Most recently,  researchers propose to train \textit{2D neural fixers} by fine-tuning diffusion models to correct artifacts at novel views by creating simulation and real pairs of held-out views~\cite{freevs,streetcrafter}, 
which yields
significant
visual improvements.
To further
improve
3D
consistency
and use for simulation purposes, subsequent methods distill the visual improvements back into the underlying 3D representation~\cite{freesim,wu2025difix3d,mudg,recondreamer,recondreamer++}.
However, despite the impressive results, these pipelines require hours of per-scene optimization and have difficulty scaling.
In addition, the distilled representations remain fragile and usually generalize only to small synthesized viewpoint shifts where the fixer performs well, with significant degradation under larger extrapolations.

Towards this goal, we present \name, a diffusion-guided \underline{g}eneralizable \underline{en}hancer for urban scene \underline{re}construction. \name takes any pre-trained 3D Gaussian representation and fixes the deficiencies within a few minutes. At the heart of \name are two modules. A one-step diffusion neural fixer predicts view-conditioned residuals at novel views, guided by geometry and appearance cues.
A generalizable 3D enhancer then updates Gaussian parameters to enforce multi-view and geometric consistency while preserving fidelity along both recorded and novel trajectories. The enhancer learns to transfer diffusion priors into iterative 3D-consistent updates by training across diverse scenes. \name{} produces stable renderings under meter-scale viewpoint shifts and lane changes, while plausibly completing unobserved or occluded regions.

Experiments on diverse urban scenes show that \name{} outperforms state-of-the-art scene reconstruction and neural fixer methods at challenging novel viewpoints while maintaining competitive performance along the recorded trajectories.
We also show that \name{} benefits various downstream tasks, including higher-quality simulation of novel maneuvers,
reduced domain gap for downstream perception tasks, and improved 3D object detection training with augmentation, unveiling the potential for robust and scalable sensor simulation.
\section{Related Work}
\label{sec:related_work}

\paragraph{Urban scene reconstruction}
Seminal works in differentiable rendering such as NeRF~\cite{nerf} and 3DGS~\cite{3dgs} have driven rapid progress in urban scene reconstruction~\cite{yang2023unisim,tonderski2024neurad,yan2024street,chen2024omnire,splatad}. These methods represent the scene as either an implicit
radiance
field or a set of 3D Gaussians which can be differentiably rendered and supervised with reconstruction losses through per-scene optimization.
To improve efficiency,
recent approaches \cite{chen2025g3r,wang2025flux4d} adopt a feed-forward paradigm, achieving substantial speedup in reconstruction.
Despite these advances, both per-scene and generalizable methods achieve high quality primarily along the training trajectories,
but their performance degrades
significantly
under large viewpoint shifts.
Our work aims to address this limitation by harnessing diffusion models to enhance the rendering quality in a 3D-consistent and efficient manner.

\paragraph{Reconstruction with generative priors}
To regularize the representation and improve visual plausibility at extrapolated views, one popular paradigm is to
incorporate
priors from large-scale generative models. A common strategy is to adapt the Score Distillation Sampling (SDS) loss~\cite{dreamfusion}, which guides the optimization of 3D representations by aligning rendered images with the gradients of a pre-trained diffusion model. While originally developed for object-level text-to-3D generation, recent methods have explored extending SDS to scene-level reconstruction~\cite{reconfusion}. In the driving domain, VEGS~\cite{vegs} adapts SDS for dynamic urban scenes and further introduce surface normal priors for regularization.
However, despite these improvements, the optimization remains unstable due to competing and sometimes noisy losses, and remains scene-specific and computationally expensive, limiting its applicability to scalable simulation.
Another line of work leverages diffusion models to directly generate 3D scenes conditioned on several input images~\cite{dreamdrive,scube,dist4d}, which typically sacrifices geometric and photometric accuracy.

\paragraph{Neural fixer of 3D scenes}
Our work is closely related to the recent advances in \textit{neural fixer} for 3D scenes, which aim to reduce artifacts such as holes, blurriness and distortions under extrapolated viewpoints.
The representative work
SplatFormer \cite{splatformer} trains a network on large-scale object-centric data~\cite{objaverseXL} to refine pre-trained 3D Gaussian representations using supervision from extreme out-of-distribution views.
However, this strategy is not applicable to urban driving scenes, where only one single pass is typically available and accurate supervision beyond the recorded trajectories is lacking.
To overcome this limitation, recent works \cite{freevs} employ generative models to generate the pseudo ground-truth at the challenging viewpoints (\textit{e.g.}, 2-3m lateral shifts), and fine-tune the 3D representation with the generated images~\cite{wu2025difix3d,freesim,recondreamer,recondreamer++}.
However, they require costly per-scene optimization, and the distilled
representations
tend to overfit to synthesized views while exhibiting noticeable artifacts under larger extrapolation.
In contrast, \name{} proposes a \textit{generalizable enhancer}, providing significant speedups and improved robustness for urban scene reconstruction.

\section{Robust Scene Reconstruction with Diffusion-guided Generalizable Enhancer}
\label{sec:method}

\begin{figure*}[t]
	\centering
	\includegraphics[width=1.0\textwidth]{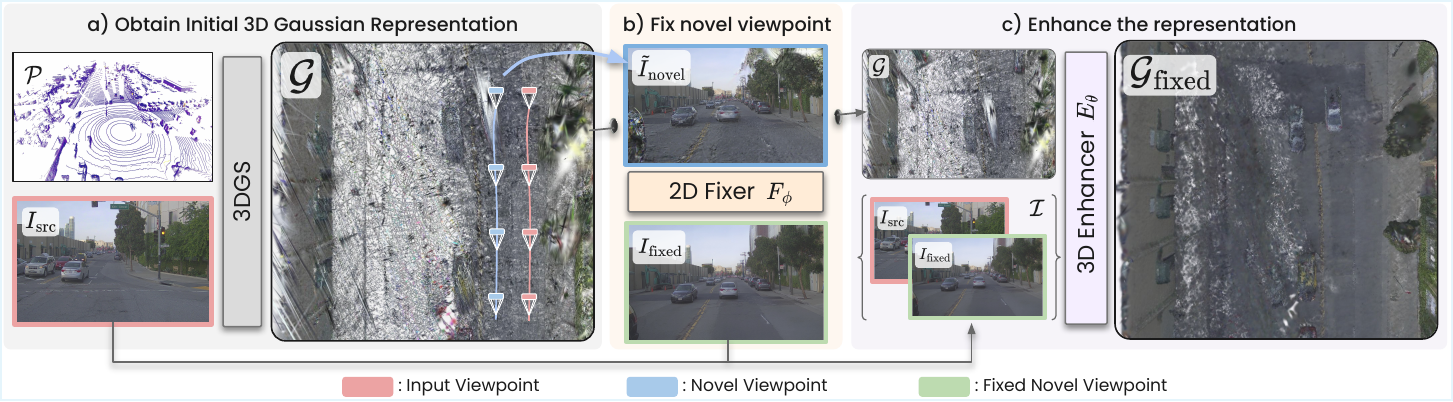}
	\vspace{-0.27in}
	\caption{
		\textbf{\name{} pipeline for urban scene reconstruction.} \name{} is composed of three steps. First, any 3DGS-based reconstruction methods are used to obtain an initial representation. Then, we render at novel viewpoint (e.g., 3m shifts) and adopt a diffusion-based neural fixer \textit{FNet} (Sec. \ref{sec:neural_fixer}) to fix the degraded artifacts. Finally, we leverage a generalizable enhancer \textit{ENet} (Sec. \ref{sec:enhancer}) that predicts per-Gaussian residuals to enhance the 3D representation.
	}
	\label{fig:pipeline}
	\vspace{-3mm}
\end{figure*}

Given multi-view camera ($\mathcal{I}_\text{src}$) and LiDAR ($\mathcal{P}$) observations from large-scale driving scenes, our goal is to reconstruct robust 3D scene representation at scale that handles large viewpoint shifts and occlusions for reliable re-simulation and downstream
evaluation.
Towards this goal, we propose a diffusion-guided generalizable 3D enhancer that improves the 3D Gaussian representation for robust rendering under challenging novel viewpoints.
We first briefly review the 3DGS-based scene representation in Sec.~\ref{sec:representation}. We then introduce two key modules: a one-step diffusion-based neural fixer (\textit{FNet}) that predicts view-conditioned residuals (Sec.~\ref{sec:neural_fixer}), and a generalizable enhancer network (\textit{ENet}) that enforces multi-view consistency by refining Gaussian attributes (Sec.~\ref{sec:enhancer}) iteratively. Finally, we show how these two modules can be integrated into a robust generalizable reconstruction pipeline (\namefull{}) for scalable urban scene simulation in Sec.~\ref{sec:reconstruction}.

\subsection{3DGS-based Scene Representation}
\label{sec:representation}
3D Gaussian Splatting (3DGS)~\cite{3dgs} represents the scene as a set of anisotropic 3D Gaussians $\mathcal{G} = \{g_i\}_{i=1}^M$ that can be differentiably rasterized in real time.
Each Gaussian $g_i = \{\boldsymbol{\mu}_i, \boldsymbol{s}_i, \boldsymbol{q}_i,
o_i,
\mathbf{c}_i\} \in \mathbb R^{14}$ consists of mean $\boldsymbol{\mu}_i \in \mathbb{R}^3$, scale vector $\mathbf{s}_i \in \mathbb{R}^{3}$,  quaternion $\boldsymbol{q}  \in \mathbb{R}^{4}$, opacity value
$o_i\in [0,1]$, 
and RGB color $\mathbf{c}_i\in[0,1]^3$.
For urban driving scenes, we decompose $\mathcal{G}$ into three subsets: a static background $\mathcal{G}_B$, dynamic actors $\mathcal{G}_A$, and a distant region $\mathcal{G}_D$ (\textit{e.g.}, far-away buildings and sky). Foreground actors are tracked across frames using 3D bounding boxes that specify their size and location. The static background and dynamic actors are initialized from aggregated LiDAR points. A fixed number of Gaussians are placed at a large distance to represent  $\mathcal{G}_D$.

Given the camera
projection matrix
$\Pi$, the 3D Gaussians are projected onto the image plane and rasterized into per-ray fragments. After depth sorting along each ray, the
color $\mathbf{C}$ of a pixel $\mathbf{p}$
is computed by front-to-back alpha compositing:
\begin{align}
\alpha_i &= o_i \exp\left(-\frac{1}{2}(\mathbf p - \hat{\boldsymbol \mu}_i)^T \hat{\mathbf \Sigma}_i^{-1} (\mathbf p - \hat{\boldsymbol \mu}_i)\right), \\
\mathbf{C}&=\sum_{i=1}^{N} w_i\,\mathbf{c}_i,\quad
w_i=\alpha_i \prod_{j<i} (1-\alpha_j),
\end{align}
where $\hat{\boldsymbol \mu}_i$ and $\hat{\mathbf \Sigma}_i$ are the projected mean and covariance of the $i$-th Gaussian, computed from its parameters ($\boldsymbol{\mu_i}$, $\boldsymbol{s_i}$, $\boldsymbol{q_i}$) and camera projection $\Pi$~\cite{3dgs}.
$\alpha_i$ is the transmittance and
$w_i$ is the
weight.
The image is rendered $\tilde{I} = f_\text{render} (\mathcal G; \Pi).$

\subsection{Enhancing NVS with 2D Neural Fixer (FNet)}
\label{sec:neural_fixer}

Although 3DGS-based reconstruction methods achieve high-quality rendering at the original or interpolated views,
they often degrade significantly at viewpoints that deviate substantially from the recorded trajectories due to overfitting to the training views and the lack of supervision in unobserved regions.
Inspired by recent works in 2D neural fixer~\cite{streetcrafter,wu2025difix3d}, we therefore learn an image-space, diffusion-based fixer that reduces
artifacts under large viewpoint shifts.

\begin{figure}[t]
	\centering
	\includegraphics[width=1.0\columnwidth]{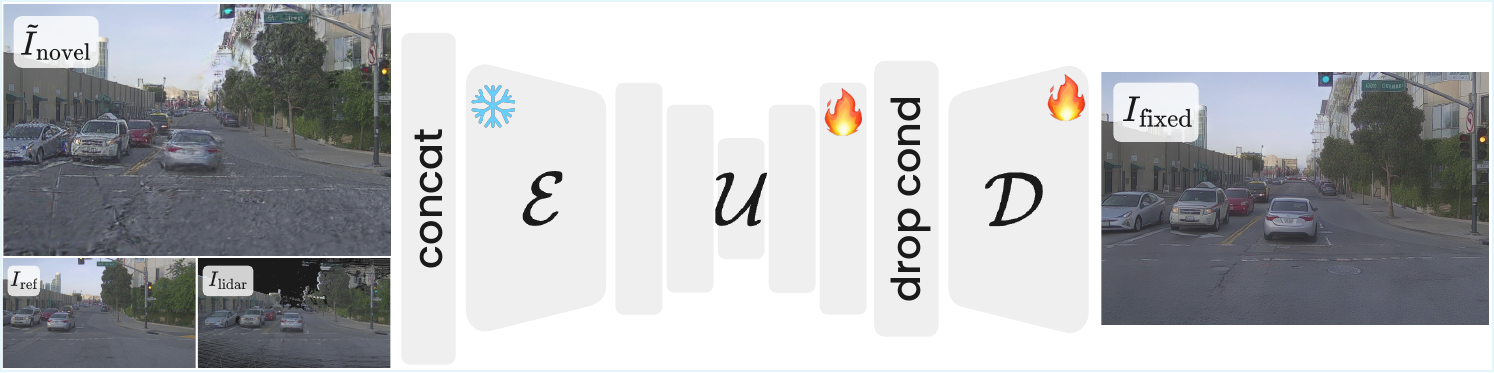}
	\vspace{-0.2in}
	\caption{
		\textbf{2D neural fixer (\textit{FNet}) overview.}
		\textit{FNet} takes a 3DGS-rendered view $\tilde{I}$, conditions on the reference image $I_{\text{ref}}$ and the rendered LiDAR map $I_{\text{lidar}}$, and produces the fixed image $I_{\text{fixed}}$. We fine-tune \textit{FNet} from the pre-trained single-step diffusion model \textit{SD-Turbo} \cite{sdturbo}.
	}
	\vspace{-0.05in}
	\label{fig:fnet}
\end{figure}

\paragraph{2D Fixer Network}

Given an image $\tilde{I}$ rendered from a pre-trained 3DGS scene, we learn a neural fixer $F_{\phi}$ to fix the rendering artifacts and obtain a more photorealistic image.
During training,  we use paired data and supervise $F_{\phi}(\tilde{I})$ to match the ground-truth image $I$.
Since $\tilde{I}$ is already close to $I$, we adopt a pre-trained single-step diffusion model \textit{SD-turbo} for efficiency.
Following~\cite{onestepdiffusion}, we encode $\tilde{I}$ into the latent space and feed it to the diffusion UNet to obtain a noisy latent. We then apply a single-step sampler to denoise the latent and decode it with the frozen VAE decoder to obtain the fixed image $I_{\text{fixed}} = F_{\phi}(\tilde{I})$. The fixer is supervised in image space using a photometric term and a perceptual term:
\begin{equation}
    \mathcal{L}_\text{fixer} = \mathcal{L}_\text{rgb} + \mathcal{L}_\text{lpips}.
\end{equation}

\paragraph{Appearance and geometry conditioning}

Although the vanilla fixer with rendered-image-only improves fidelity, it still produces blurry or hallucinatory content and, as an image-based diffusion model, lacks temporal consistency.
To improve fidelity and consistency, we augment the 2D fixer with additional appearance and geometry conditioning.
In particular, we take as the reference view $I_{\text{ref}}$ the training image whose camera pose is closest to the target view, and we render an accumulated LiDAR map $I_{\text{lidar}}$ (\textit{i.e.}, a 3DGS rendering of aggregated, colored LiDAR points at the target view) to provide explicit geometric cues (Fig.~\ref{fig:fnet}).
Formally, let $\mathcal{E}$ and $\mathcal D$ denote the VAE encoder and decoder, $\tau$  be the noise timestep with schedule $\sigma_\tau$, and $\mathcal{U}$ be the single-step diffusion UNet, we have
\begin{align}
	\mathbf{z}_{\text{render}} &= \mathcal{E}(\tilde{I}), \quad
	\mathbf{z}_{\text{ref}} = \mathcal{E}(I_{\text{ref}}), \quad
	\mathbf{z}_{\text{lidar}} = \mathcal{E}(I_{\text{lidar}}),\\
	\mathbf{z}_\tau &= \mathbf{z}_{\text{render}} + \sigma_\tau \boldsymbol{\epsilon}, \quad \boldsymbol{\epsilon}\sim\mathcal{N}(\mathbf{0},\mathbf{I}),\\
	\tilde{\boldsymbol{\epsilon}} &= \mathcal{U}(\mathbf{z}_{\tau}, \tau | [\mathbf{z}_{\text{ref}}, \mathbf{z}_{\text{lidar}}]),\\
	\tilde{\mathbf{z}} &= \mathbf{z}_\tau - \sigma_\tau \tilde{\boldsymbol{\epsilon}}, \quad
	I_{\text{fixed}} = \mathcal{D}(\tilde{\mathbf{z}}),
\end{align}
where we encode $\tilde{I}$, $I_{\text{ref}}$, and $I_{\text{lidar}}$ with a shared VAE encoder and feed their latents $\mathbf z$ through the UNet.
Inspired by \cite{wu2025difix3d,wonder3d}, we stack latents along the view axis, and each UNet block applies a lightweight condition-mixing self-attention to share information across views before restoring the layout. This conditions denoising on appearance cues from the reference and geometric cues from LiDAR, improving view consistency and reducing
blurriness,
hallucination, and flickering under large viewpoint shifts. Fig.~\ref{fig:fnet} shows the overview of \textit{FNet}.

\paragraph{Implementation details}
We initialize the VAE ($\mathcal{E}, \mathcal{D}$) and UNet ($\mathcal{U}$) from pre-trained \textit{SD-Turbo}~\cite{sdturbo} and fine-tune $\mathcal{D}$ and $\mathcal{U}$ (with LoRA~\cite{lora}).
We remove the CLIP cross-attention layers and set the noise timestep to $\tau = 200$ following \cite{wu2025difix3d}.
We train FNet at the resolution of $720\times1280$ for 20k steps using AdamW with a batch size of 8.

\subsection{Generalizable 3D Enhancer Network (ENet)}
\label{sec:enhancer}

While \textit{FNet} improves novel-view renderings, its per-frame runtime limits real-time use, and image-space corrections alone do not guarantee multi-view consistency. For fast simulation, the improvements must reside in the 3D representation.
Prior work address this by distilling corrected images back into 3D on a per-scene basis, which takes hours and often yields models that generalize only to small synthesized shifts with noticeable degradation under larger extrapolations.
Motivated by this gap, we propose a generalizable 3D enhancer $E_{\theta}$ that updates 3DGS parameters in an iterative manner. Trained across diverse scenes, the 3D enhancer transfers the knowledge of the 2D fixer into a 3D-consistent and robust representation within a few feed-forward steps.

\paragraph{3D Enhancer Network}

Given a pre-trained 3DGS scene $\mathcal{G}$ and the fixer $F_\theta$, we train a generalizable enhancer $E_\theta$ that predicts per-Gaussian residuals to update ${\mathcal{G}}$ into a higher-fidelity representation ${\mathcal{G}_{\text{fixed}}}$.
Specifically, conditioned on camera poses $\Pi$ and images $\mathcal{I} = \{\mathcal{I}_\mathrm{src}, \mathcal{I}_\text{fixed}\}$ (a mixture of ground-truth frames $I$ and fixer outputs $I_{\text{fixed}}=F_\phi(\tilde{I})$), the enhancer network outputs
\begin{align}
\Delta \mathcal{G} = E_\theta(\mathcal{G}; \mathcal{I},\Pi)
= \{\Delta\boldsymbol{\mu}_i, \Delta\boldsymbol{s}_i, \Delta\boldsymbol{q}_i, \Delta o_i, \Delta\boldsymbol{c}_i\}_{i=1}^M.
\end{align}
To obtain the final representation, we apply the residuals to the initial 3D Gaussians:
\begin{align}
{\mathcal{G}_\mathrm{fixed}} = \mathcal{G} + \Delta \mathcal{G}  = \{ \Delta g_i + g_i \}_{i=1}^{M}.
\end{align}
Let $\tilde{I}_{\text{src}}$ and $\tilde{I}_{\text{novel}}$ denote renders from the current 3DGS at recorded and extrapolated viewpoints.
We train $E_\theta$ across many scenes by minimizing the following training objective:
\begin{align}
    \label{eq:rendering}
    \mathcal{L} &= \mathcal{L}_\text{src} (\tilde{I}_{\text{src}}, I) + \lambda_\text{novel}\mathcal{L}_\text{novel} (\tilde{I}_\text{novel}, I_\text{fixed}), \\
    \mathcal{L}_\text{src} &= \mathcal{L}_\text{rgb} + \lambda_\text{lpips}\mathcal{L}_\text{lpips} + \lambda_\text{ssim}\mathcal{L}_\text{ssim} + \lambda_\text{depth}\mathcal{L}_\text{depth}, \\
    \mathcal{L}_\text{novel} &= \mathcal{L}_\text{rgb\_novel} + \lambda_\text{lpips}\mathcal{L}_\text{lpips\_novel} + \lambda_\text{ssim}\mathcal{L}_\text{ssim\_novel},
\end{align}
where the source-view terms compare $\tilde{I}_{\text{src}}$ with ground-truth images $I$ to preserve fidelity along the recorded trajectories, while the novel-view terms compare $\tilde{I}_{\text{novel}}$ with fixer targets $I_{\text{fixed}}$ to improve robustness under challenging viewpoints.

\paragraph{Iterative refinement}

Inspired by G3R~\cite{chen2025g3r}, we unroll the enhancer for $T$ iterations instead of using a single forward pass. At iteration $t\in\{0,\dots,T\!-\!1\}$, the network $E_{\theta}$ takes the current Gaussians $\mathcal{G}_t$ (with $\mathcal{G}_0=\mathcal{G}$) and per-point gradients $\nabla\mathcal{G}_t$ as guidance features, computed by backpropagating the loss in Eq.~\ref{eq:rendering} with respect to the Gaussian parameters. Given the predicted residuals $\Delta\mathcal{G}_t$, we then update the representation to obtain $\mathcal{G}_{t+1}$. The weights of $E_{\theta}$ are shared across iterations and the enhanced Gaussians are $\mathcal{G}_\text{fixed} = \mathcal{G}_T$.
This iterative refinement (also known as learned optimization) substantially improves visual quality under extrapolated viewpoints.

\paragraph{Implementation details}
We use Sparse UNet as the 3D enhancer network.
We set $T=12$, and train the model for 2k iterations with a batch size of 8.
 We set $\lambda_\text{rgb} = 0.8$, $\lambda_\text{lpips} = 0.2$, $\lambda_\text{ssim} = 0.2$, $\lambda_\text{depth} = 0.01$, $\lambda_\text{novel} = 0.5$.
We split each training sequence into 20-frame non-overlapping chunks, and optimize the 3DGS scene for each chunk to obtain $\mathcal{G}$ for training \textit{ENet}.

\begin{figure}[t]
	\centering
	\includegraphics[width=1.0\columnwidth]{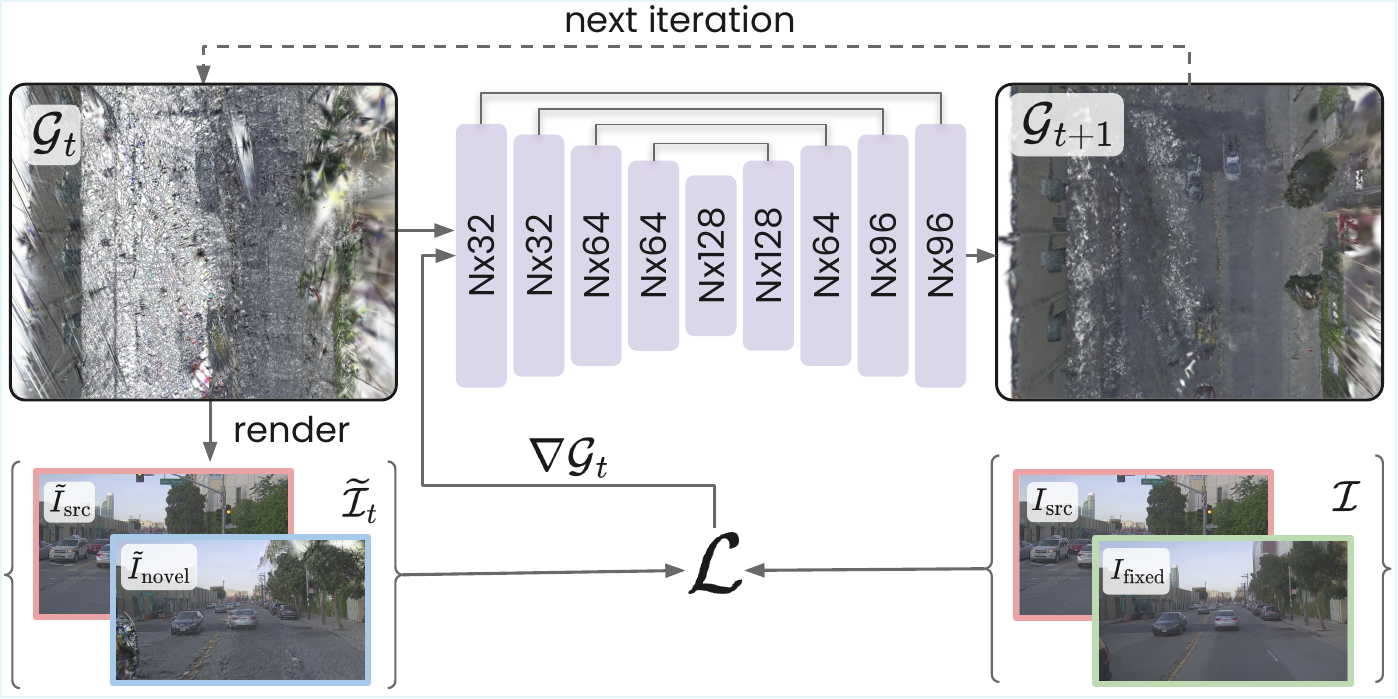}
	\vspace{-0.2in}
	\caption{\textbf{Generalizable 3D enhancer
	 	(\textit{ENet}) overview.}
		\textit{ENet} iteratively refines a 3DGS scene using rendering-guided gradients.
		At iteration $t$, \textit{ENet} takes the current 3D Gaussians $\mathcal{G}_t$ and
		per-Gaussian gradients $\nabla\mathcal{G}_t$ (from rendering loss) and predicts residuals $\Delta\mathcal{G}_t$
		to
		update the scene to $\mathcal{G}_{t+1}$.
		Source and novel views are compared with ground-truth $I$ and fixed targets $I_{\text{fixed}}$
		to
		compute
		losses $\mathcal{L}_{\text{src}}(\tilde{I}_{\text{src}},I)$ and $\mathcal{L}_{\text{novel}}(\tilde{I}_{\text{novel}},I_{\text{fixed}})$, whose backprop gives $\nabla\mathcal{G}_{t+1}$.
		Unrolling $T$ steps yields the enhanced scene $\mathcal{G}_{\text{fixed}}$.
	}
	\label{fig:enet}
	\vspace{-0.1in}
\end{figure}

 \subsection{\name{} (Enhancer) and \namefull{} (Reconstructor)}
\label{sec:reconstruction}

We now describe how to integrate the neural fixer \textit{FNet} and the 3D enhancer \textit{ENet} into a unified framework, \name{}, for robust urban scene reconstruction. As shown in Fig.~\ref{fig:pipeline}, starting from an initial 3DGS representation $\mathcal{G}$, we first render novel viewpoints and correct artifacts with \textit{FNet}. These fixed images $\mathcal I_\mathrm{fixed}$ are then used by \textit{ENet} together with source-view real images $\mathcal I_\mathrm{src}$ to refine the underlying Gaussian representation, distilling the 2D corrections back into 3D. Formally, we have
\begin{align}
	\mathcal{I}_\text{fixed} &=  F_{\phi} (f_\text{render} (\mathcal{G}; \Pi_\text{novel})) \\
	\mathcal{G}_\textrm{fixed} &= E_{\theta} (\mathcal{G}, \{\mathcal{I}_\text{src}, \mathcal{I}_\text{fixed}\}; \{\Pi_\text{src}, \Pi_\text{novel} \}).
\end{align}

Although \textit{ENet} is designed to enhance existing 3DGS scenes, its formulation as a learned optimizer makes it naturally amenable to inducing a 3DGS representation directly from data.
{We therefore adapt it as a robust and efficient}
generalizable reconstruction module that predicts scenes from raw sensory inputs, which we refer to as \textit{GNet}.
To further improve robustness under extrapolation, we include \textit{FNet}-generated images as auxiliary supervision during training with a small weight. This preserves fidelity along recorded trajectories while strengthening stability at challenging novel viewpoints.
Benefiting from the 2D fixer priors,
\textit{GNet} provides strong standalone reconstruction and achieves superior robustness compared to existing
methods.
We obtain \textit{GNet} by fine-tuning the enhancer $E_{\theta}$ with the rendering objective in Eq.~\ref{eq:rendering} ($\lambda_{\text{novel}}{=}0.1$).
We unroll $T{=}24$ iterations to increase reconstruction capacity.

Finally, we show that combining \textit{GNet} (generalizable reconstruction), \textit{FNet} (2D fixer), and \textit{ENet} (3D enhancer) yields a more robust and scalable pipeline, \namefull{} (\textit{GNet} $\rightarrow$ \textit{FNet} $\rightarrow$ \textit{ENet}), for urban scene reconstruction.
Specifically, \textit{GNet} reconstructs the base scene representation from sensory data; \textit{FNet} corrects artifacts at novel-view rendering; and \textit{ENet} distills these corrections back into the 3D representation.

\section{Experiments}
\label{sec:experiment}

\begin{table*}[t]
	\centering
	\caption{Comparison to state-of-the-art reconstruction methods on interpolated and extrapolated views.}
	\vspace{-0.05in}
	\label{tab:main_table}
	\resizebox{0.95\textwidth}{!}{
		\begin{tabular}{lccccccccc}
			\toprule
			\multirow{2}{*}{\textbf{Methods}}
			& \multicolumn{2}{c}{\textbf{\textit{Interpolation}}}
			& \multicolumn{3}{c}{\textbf{\textit{Extrapolation} (Moderate)}}
			& \multicolumn{2}{c}{\textbf{\textit{Extrapolation} (Hard)}}
			& \multirow{2}{*}{\textbf{Recon time}} \\
			\cmidrule(lr){2-3} \cmidrule(lr){4-6} \cmidrule(lr){7-8}
			& PSNR$\uparrow$ & SSIM$\uparrow$
			& FID@1m$\downarrow$ & FID@2m$\downarrow$ & FID@3m$\downarrow$
			& FID@4m$\downarrow$ & FID@5m$\downarrow$ & Minute$\downarrow$\\
			\midrule
			\multicolumn{9}{l}{\textit{\textbf{Standalone reconstruction}}} \\
			3DGS \cite{3dgs} & 23.45 & {0.707} & 82.63 & 128.10 & 169.69 & 205.09 & 231.70 & 41.90 \\
			StreetGS \cite{yan2024street} & 23.14 & 0.693 & \textbf{68.99} & {97.05} & {127.06} & {153.76} & {176.43} & 47.62 \\
			SplatAD \cite{splatad} & \textbf{24.93} & \textbf{0.768} & 84.21 & 122.56 & 160.43 & 188.00 & 210.24 & 113.62 \\
			G3R \cite{chen2025g3r} & 23.28 & 0.673 & 89.75 & 114.94 & 147.50 & 174.64 & 191.33 & \textbf{0.90} \\
			Ours (\textit{GNet}) & {23.56} & 0.689 & {70.04} & \textbf{86.43} & \textbf{106.07} & \textbf{124.65} & \textbf{138.30} & {0.96} \\
			\midrule
			\multicolumn{9}{l}{\textit{\textbf{Reconstruction with neural fixers}}} \\
			StreetCrafter \cite{streetcrafter} & {23.33} & 0.690 & \textbf{59.44} & {81.14} & {97.09} & {118.94} & {141.14} & 127.33 \\
			Difix3D \cite{wu2025difix3d} & \textbf{23.34} & \textbf{0.705} & {60.29} & 83.35 & 102.16 & 137.80 & 167.28 & {35.32} \\
			Ours (\namefull) & 23.28 & {0.692} & 60.69 & \textbf{74.50} & \textbf{88.04} & \textbf{102.73} & \textbf{114.19} & \textbf{2.77} \\
			\bottomrule
		\end{tabular}
	}
\end{table*}

\begin{figure*}[t]
	\centering
	\includegraphics[width=1.0\textwidth]{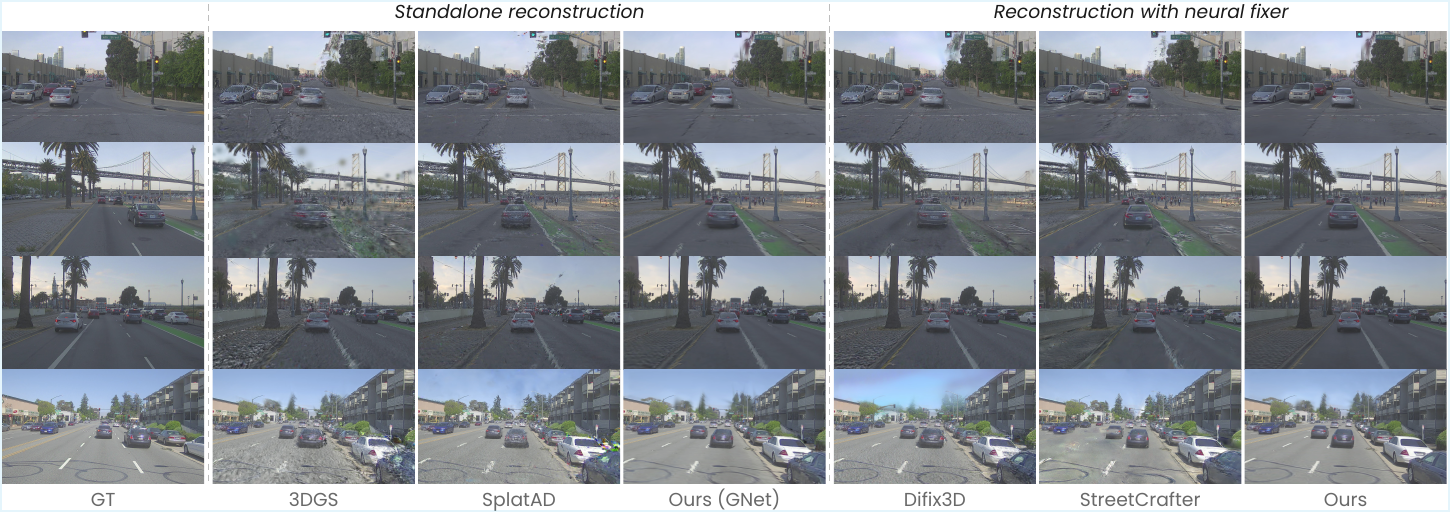}
	\vspace{-0.25in}
	\caption{
		Qualitative comparison to state-of-the-art neural
		reconstruction methods under large extrapolation.
		Our method yields higher realism, fewer artifacts.
	}
	\label{fig:recon_viz}
	\vspace{-2mm}
\end{figure*}

We evaluate against state-of-the-art  (SoTA) urban scene reconstruction approaches and per-scene optimization with neural fixers. The performance is measured on both recorded trajectories and extrapolated viewpoints (\textit{i.e.,} lateral shifts). We then show that our generalizable enhancer \name plugs into different 3DGS-based methods in a zero-shot manner, demonstrating its versatility and robustness.
Finally, we showcase \namefull{} benefits  various downstream tasks including simulation, perception evaluation and augmented training.

\subsection{Experiment Details}
\paragraph{Experiment setup}
We evaluate on PandaSet~\cite{pandaset}, a self-driving dataset with diverse, large-scale urban scenes.
PandaSet contains 103 sequences captured by six 1080p cameras and a 64-beam LiDAR at 10Hz. We follow the split of~ \cite{chen2025g3r}, using 93 sequences for training, and 10 for testing.
We assess performance on both in-trajectory (\textit{interpolation}) and out-of-trajectory (\textit{extrapolation}) views. In all experiments, we subsample every fourth frame as input (25\% of views) to reconstruct the scene representation and use the remaining 75\% for interpolation evaluation. For extrapolation, we synthesize lateral shifts of 1–5$m$ (1-3$m$: \textit{moderate}; 4–5$m$: \textit{hard}) and report FID~\cite{fid}. For methods with neural fixers, we apply the fixers only at 3$m$, where prior work reports reasonable fidelity without pronounced consistency issues or hallucination~\cite{streetcrafter} and evaluate up to 5$m$ to test the robustness under larger extrapolations.

\paragraph{Baselines}
We compare \name{} against SoTA urban scene reconstruction in two settings: (1) \textit{standalone reconstruction}, including per-scene methods 3DGS~\cite{3dgs}, StreetGS~\cite{yan2024street} and SplatAD~\cite{splatad}, as well as the generalizable method G3R~\cite{chen2025g3r}; and (2) r\textit{econstruction with neural fixers}, including StreetCrafter~\cite{streetcrafter} and Difix3D~\cite{wu2025difix3d}, where diffusion-based 2D image fixers refine novel views and the improvements are distilled back into the 3D representation through per-scene optimization.
We train Difix3D on PandaSet following the official repository\footnote{\href{https://github.com/nv-tlabs/Difix3D}{Difix3D official repository}}, and use the publicly released model checkpoint trained on PandaSet for StreetCrafter\footnote{\href{https://drive.google.com/file/d/1Qtdkm0wvIUSMWQMVldd-d16rHZsNFFt1/view}{StreetCrafter official model weights}}.

\subsection{Experimental Results}

\paragraph{Comparison to SoTA reconstruction methods}

Table~\ref{tab:main_table} reports the quantitative results. When comparing to standalone reconstruction approaches, \textit{GNet} surpasses all baselines by a large margin in FID across nearly all lateral offsets while remaining competitive under original trajectories. This indicates that our generalizable enhancer can be efficiently adapted as a standalone reconstruction method, producing high-quality, robust 3D representations.
In the \textit{reconstruction with neural fixers} setting, our full pipeline, \name{}, which integrates reconstruction with a diffusion-based 2D fixer and a generalizable 3D enhancer, achieves the best performance at challenging views while being substantially more efficient ($100\times$). It outperforms SoTA methods StreetCrafter and Difix3D especially under large extrapolations
(\textit{hard}). Qualitative results in Fig.~\ref{fig:recon_viz} further show more complete reconstructions and fewer view-dependent artifacts for \name{} at extreme viewpoints.

\paragraph{Generalizable enhancements on varied 3DGS-based methods}
We then demonstrate that our generalizable enhancer \name{} is generic and can easily plug into different 3DGS-based methods to enhance the representation. As shown in Table~\ref{tab:different_rep}, our enhancer, when trained on vanilla 3DGS, is able to
correct the deficiencies in representation, and significantly boost the rendering quality at both original trajectory and
extrapolated views.
We also apply the pre-trained enhancer to directly refine the 3DGS representation produced by G3R in a zero-shot manner, achieving substantial improvements across novel viewpoints without retraining or altering the backbone, which demonstrates its versatility and robustness.
A brief fine-tuning stage on the G3R representation further brings additional improvements.

\begin{figure}[t]
	\centering
	\includegraphics[width=1.0\columnwidth]{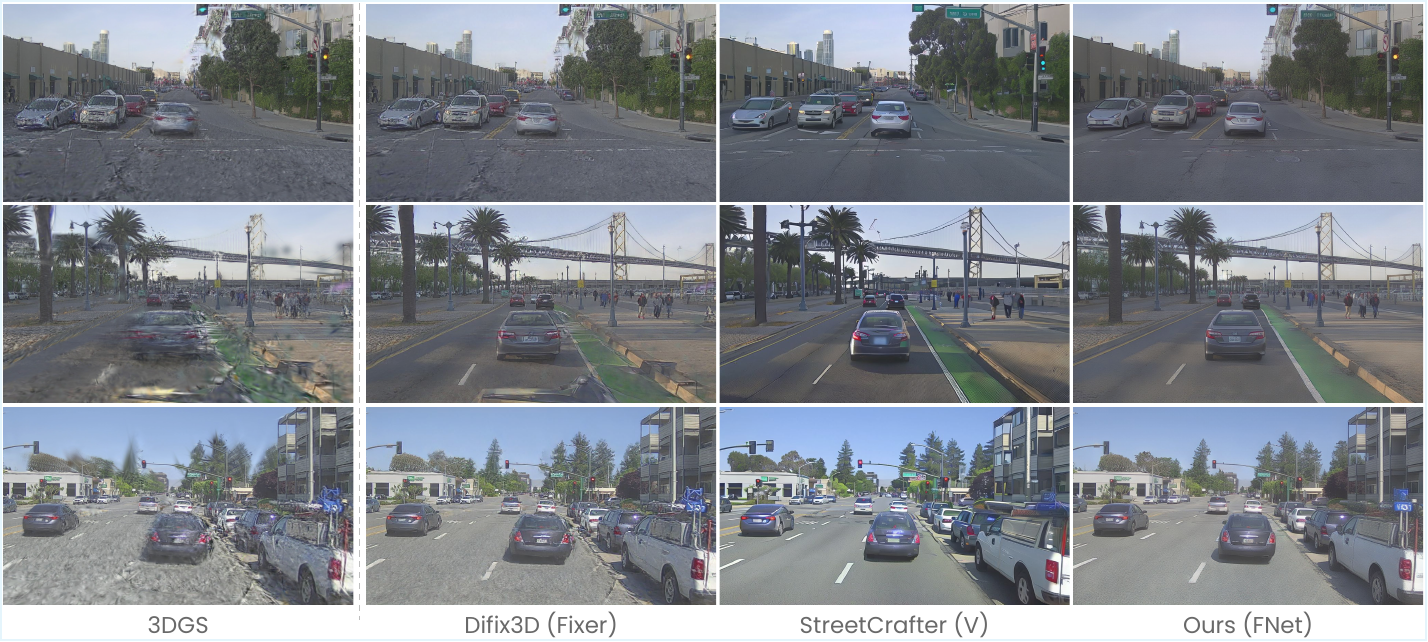}
	\vspace{-0.25in}
	\caption{
		Qualitative comparison to state-of-the-art 2D neural fixers.
	}
	\label{fig:fixer_viz}
\end{figure}

\paragraph{2D neural fixer comparison}

Table~\ref{tab:fixer_comparison} compares our 2D neural fixer (\textit{FNet}) with the SoTA baselines. StreetCrafter-V~\cite{streetcrafter}, the fine-tuned video diffusion model in StreetCrafter, conditions on the reference view (\textit{ref}) and the 3DGS rendering at the target view from colored LiDAR points (\textit{lidar}). Difix3D~\cite{wu2025difix3d} conditions on the rendered source image (\textit{src}) and the reference image. In contrast, \textit{FNet} leverages all three signals: \textit{src} + \textit{ref} + \textit{lidar}. It achieves the lowest FID across all lateral shifts, demonstrating the advantage of pairing appearance cues with an explicit geometry-backed render. Qualitatively results in Fig.~\ref{fig:fixer_viz} show \textit{FNet} produces less stylized, more photorealistic results than StreetCrafter-V and maintains stronger geometric consistency than Difix3D.

\begin{table}[t]
	\centering
	\caption{3D enhancer plugs into different 3DGS-based methods.}
	\vspace{-0.05in}
	\label{tab:different_rep}
	\resizebox{\columnwidth}{!}{
		\begin{tabular}{lcccc}
			\toprule
			\textbf{Methods} & FID@0m$\downarrow$
			& FID@1m$\downarrow$ & FID@2m$\downarrow$ & FID@3m$\downarrow$\\
			\midrule
			3DGS \cite{3dgs} & 61.74 & 82.45 & 117.05 & 154.21  \\
			+ \name & \textbf{57.32} & \textbf{69.62} & \textbf{85.02} & \textbf{99.69} \\
			\midrule
			G3R \cite{chen2025g3r} & 70.62 & 80.48 & 104.75 & 132.46 \\
			+ \name (zero-shot) & {65.75} & {74.86} & {89.22}  & {100.67}\\
			+ \name (fine-tune) & \textbf{55.31} & \textbf{65.90} & \textbf{79.41} & \textbf{91.12} \\
			\bottomrule
		\end{tabular}
	}
\end{table}

\begin{table}[t]
	\centering
	\caption{Comparison to state-of-the-art 2D neural fixers.}
	\vspace{-0.07in}
	\label{tab:fixer_comparison}
	\resizebox{\columnwidth}{!}{
		\begin{tabular}{lccccccccc}
			\toprule
			\multirow{2}{*}{\textbf{Methods}}
			& \multicolumn{3}{c}{\textbf{Input}}
			& \multicolumn{3}{c}{\textbf{FID} $\downarrow$}
			& \multirow{2}{*}{\textbf{Inference}}  \\
			\cmidrule(lr){2-4} \cmidrule(lr){5-7}
			& \textit{src} & \textit{ref} & \textit{lidar}
			& @1m & @2m & @3m
			&  Time$\downarrow$ \\
			\midrule
			StreetCrafter (V) &  & \cmark & \cmark & 65.60 & 80.46 & 92.98 & 15.26 s/frame \\
			Difix3D (Fixer) & \cmark & \cmark & & {59.07} & {75.33} & {91.11} & \textbf{0.57 s/frame} \\
			Ours (\textit{FNet}) & \cmark & \cmark & \cmark & \textbf{50.12} & \textbf{65.55} & \textbf{80.27} & {0.83 s/frame}\\
			\bottomrule
		\end{tabular}
	}
\end{table}

\begin{table}[t]
	\centering
	\caption{Ablation study on \name{} components.}
	\vspace{-0.07in}
	\label{tab:ablation}
	\resizebox{\columnwidth}{!}{
		\begin{tabular}{lccccc}
			\toprule
			\multirow{2}{*}{\textbf{Methods}} & \multicolumn{5}{c}{\textbf{Extrapolation FID} $\downarrow$ } \\
			\cmidrule(lr){2-6}
			& @1m & @2m & @3m & @4m & @5m\\
			\midrule
			\textit{GNet} & 70.04 & 86.43 & 106.07 & 124.65 & 138.30 \\
			\textit{GNet}$\rightarrow$\textit{FNet}$\rightarrow$\textit{GNet} & {64.56} & {78.58} & {91.20} & {105.25} & {117.43} \\
			\textit{GNet}$\rightarrow$\textit{FNet}$\rightarrow$\textit{ENet} & \textbf{60.69} & \textbf{74.50} & \textbf{88.04} & \textbf{102.73} & \textbf{114.19} \\
			\bottomrule
		\end{tabular}
	}
\end{table}

\begin{table}[htbp!]
	\centering
	\caption{Runtime analysis on Difix3D and \name{} (in
		minutes).}
	\vspace{-0.07in}
	\label{tab:runtime}
	\resizebox{\columnwidth}{!}{
		\begin{tabular}{lcccc}
			\toprule
			\textbf{Methods} & Reconstruction & Neural Fixer & Distillation & Total \\
			\midrule
			Difix3D \cite{wu2025difix3d} & 12.43 & \textbf{0.383} & 22.50 & 35.32\\
			\namefull & \textbf{0.967} & 0.550 & \textbf{1.25} & \textbf{2.77} \\
			\bottomrule
		\end{tabular}
	}
\end{table}

\begin{table}[htbp!]
	\centering
	\caption{Re-simulation evaluation with different behaviors.}
	\vspace{-0.07in}
	\label{tab:sim_eval}
	\resizebox{\columnwidth}{!}{
		\begin{tabular}{lccccc}
			\toprule
			\multirow{1}{*}{\textbf{Methods}} & Brake & Accelerate & Change Lane & Swerve \\
			\midrule
			3DGS \cite{3dgs} & 225.40 & 234.65 & 132.28 & 129.90\\
			Difix3D \cite{wu2025difix3d} & {182.40} &	{192.08} & {84.48} & {85.99}\\
			\namefull & \textbf{152.38} & \textbf{143.03} & \textbf{77.69} & \textbf{78.49} \\
			\bottomrule
		\end{tabular}
	}
\end{table}

\paragraph{Ablation study}

We decompose \name{} into \textit{GNet} (base reconstruction), \textit{FNet} (2D image fixer), and \textit{ENet} (3D enhancer that distills fixes back into the scene representation) and quantify their contributions in Table~\ref{tab:ablation}. Correcting \textit{GNet} artifacts with \textit{FNet} and then rerunning reconstruction consistently lowers extrapolation FID, validating the effectiveness of the 2D fixer at novel views. Replacing the final \textit{GNet} pass with \textit{ENet} yields further gains in quality and efficiency, indicating distilling 2D fixes into 3D is more effective than rerunning reconstruction.

\paragraph{Runtime analysis}
We provide a detailed runtime analysis compared to Difix3D in Table \ref{tab:runtime}.
On average, \namefull{} reconstructs a scene in 2.8 minutes, yielding a $10\times$ speedup over Difix3D (35.3 minutes).
The speedup mainly comes from the distillation process, where Difix3D relies on costly per-scene reconstruction whereas our approach leverages a efficient generalizable 3D enhancer. Notably, our 3D enhancer \name{} (\textit{FNet} $\rightarrow$ \textit{ENet}) takes only 1.8 minutes to correct deficiencies in a 3D scene.

\subsection{Downstream Applications}

\begin{figure}[t]
	\centering
	\includegraphics[width=1.0\columnwidth]{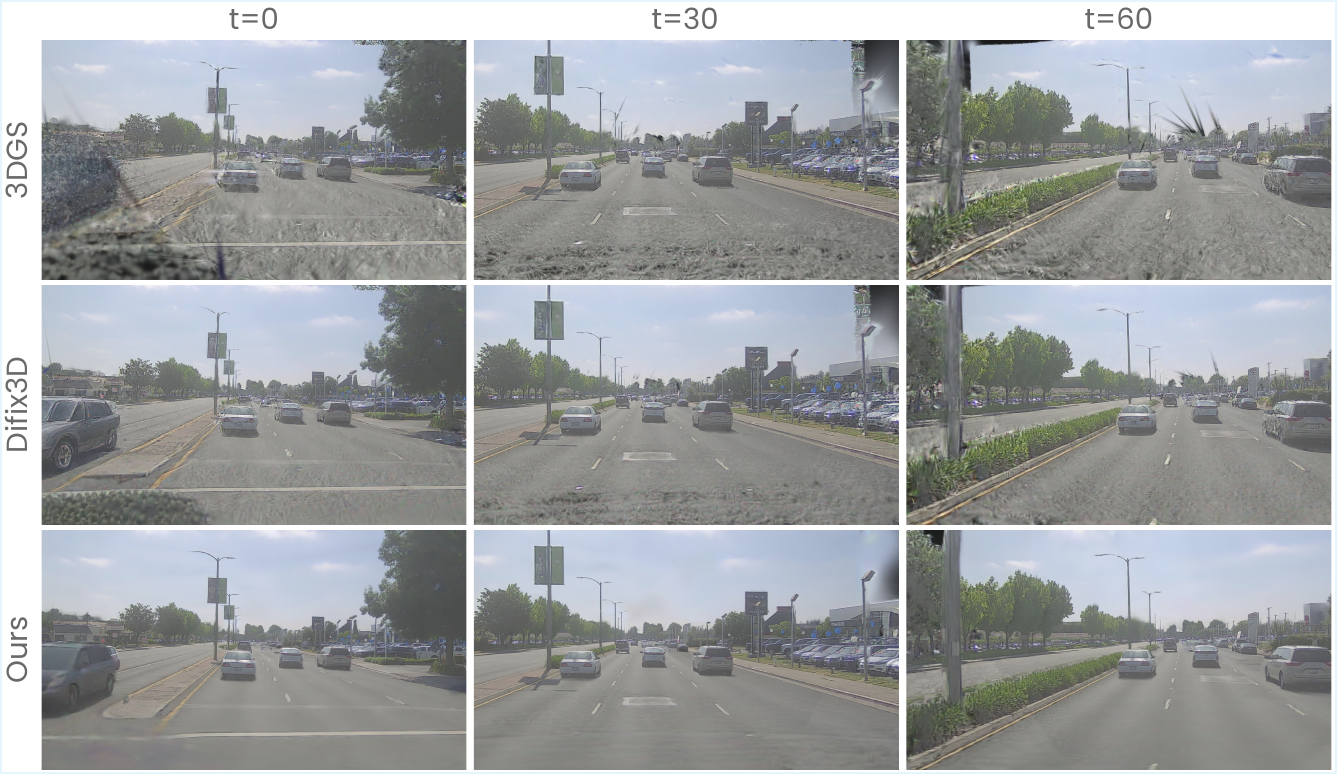}
	\vspace{-0.25in}
	\caption{
		Qualitative results on re-simulation in \textit{swerving} behavior.
	}
	\vspace{-0.1in}
	\label{fig:swerve}
\end{figure}

\begin{figure}[t]
	\centering
	\includegraphics[width=1.0\columnwidth]{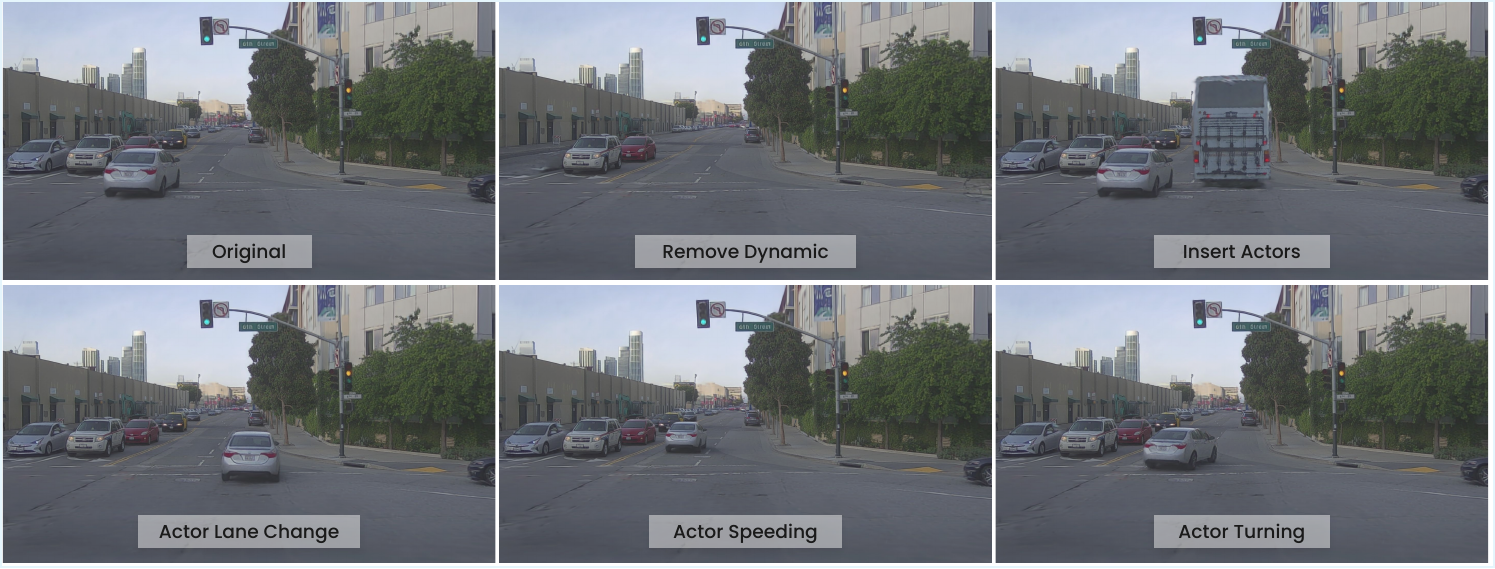}
	\vspace{-0.25in}
	\caption{
		\namefull{} can support diverse variants for reactive log replay, such as dynamic actor removals, actors insertions, and actors manipulation.
	}
	\vspace{-0.1in}
	\label{fig:manipulation}
\end{figure}

\paragraph{Realistic re-simulation with different behaviors}
To test whether more robust reconstruction benefits downstream simulation, we emulate open-loop re-simulation by branching from the recorded trajectory and rendering along perturbed ego paths. We consider four behaviors: \emph{braking}, \emph{acceleration}, \emph{lane change}, and \emph{swerving} (changing to another lane and then back). Each rollout starts from a lateral offset of $3$\,m, and all synthetic scenarios are manually vetted for plausibility. We report image quality (FID) against baselines. As shown in Tab. \ref{tab:sim_eval} and Fig. \ref{fig:swerve}, \namefull{} yields lower FID and provides higher quality rendering for all behaviors compared to baselines, showing its performance in downstream simulation. Moreover, Fig.~\ref{fig:manipulation} shows that \namefull{} is able to support diverse variants for reactive log replay, such as dynamic actor removals, actors insertions, and actors manipulation, with high-quality, realistic rendering, which is desired for safety-critical evaluation of autonomy system.

\paragraph{Domain gap evaluation for perception}

\begin{figure}[t]
	\centering
	\includegraphics[width=1.0\columnwidth]{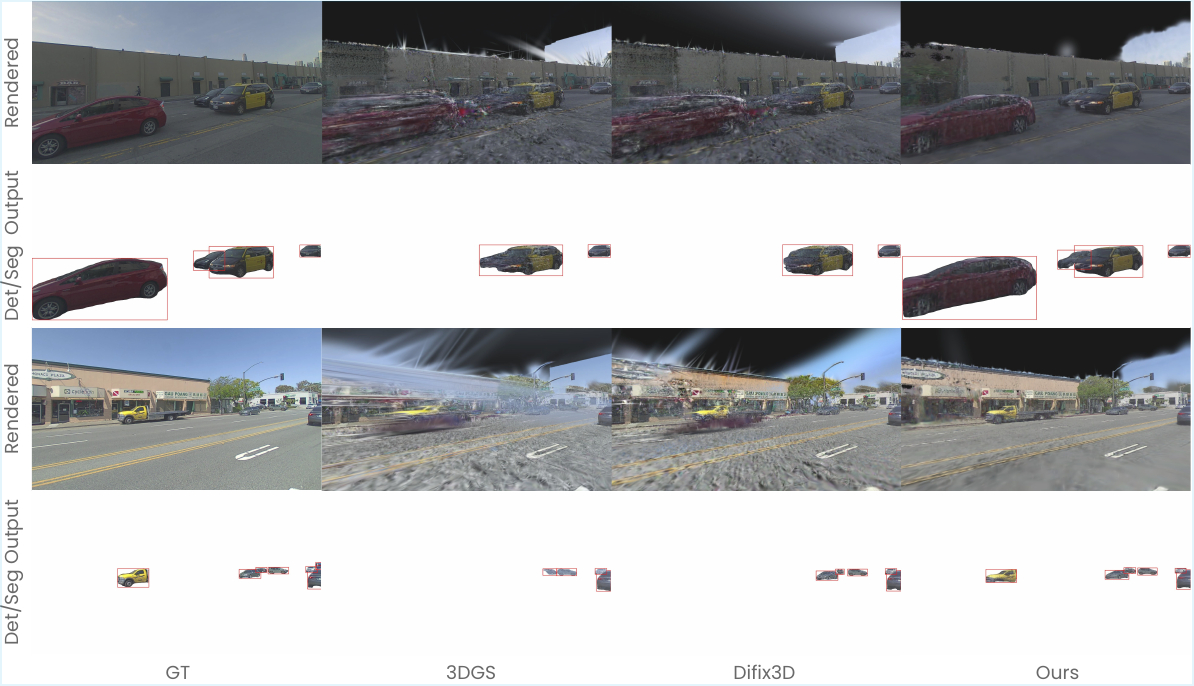}
	\vspace{-0.25in}
	\caption{
		\namefull{} shows minimal detection and segmentation domain gap.
	}
	\vspace{-0.07in}
	\label{fig:domain_gap}
\end{figure}

To evaluate how well the reconstruction methods can be used to test existing perception systems at challenging viewpoints, we measure domain gap via the \emph{perception-agreement} metric under novel camera synthesis.
Specifically, each scene is reconstructed using only the \textit{front} camera, then rendered from \textit{front-left} camera viewpoints. We run off-the-shelf object detection and instance segmentation models \cite{wu2019detectron2} on real and corresponding simulated renders. For each matched instance, we compute the AP, Recall, and IoU between predictions on real vs.\ simulated images (boxes and masks) and average over instances. This cross-camera agreement reflects how faithfully simulation preserves cues used by perception models under viewpoint transfer. As shown in Table~\ref{tab:domain_gap}, \namefull{} achieves the highest agreement for both detection and instance segmentation when compared to other baselines. Fig~\ref{fig:domain_gap} also illustrates these metrics, indicating \namefull's minimal domain gap and robust reconstruction.

\begin{table}[htbp!]
	\centering
	\caption{Downstream domain gap evaluation.}
	\vspace{-0.07in}
	\label{tab:domain_gap}
	\resizebox{\columnwidth}{!}{
		\begin{tabular}{lcccccc}
			\toprule
			\multirow{2}{*}{\textbf{Methods}}
			& \multicolumn{3}{c}{\textbf{Detection}}
			& \multicolumn{3}{c}{\textbf{Segmentation}} \\
			\cmidrule(lr){2-4} \cmidrule(lr){5-7}
			& AP $\uparrow$ & Recall $\uparrow$ & IoU $\uparrow$
			& AP $\uparrow$ & Recall $\uparrow$ & IoU $\uparrow$\\
			\midrule
			3DGS \cite{3dgs} & 0.560 & 0.376 & 0.505 & 0.558 & 0.375 & 0.501 \\
			Difix3D \cite{wu2025difix3d} & {0.670} & {0.434} & {0.611} & {0.670} & {0.434} & {0.598} \\
			\namefull & \textbf{0.785} & \textbf{0.607} & \textbf{0.728} & \textbf{0.768} & \textbf{0.596} & \textbf{0.723} \\
			\bottomrule
		\end{tabular}
	}
\end{table}

\paragraph{3D object detection with simulated data}
\begin{table}[htbp!]
	\centering
	\caption{Downstream training with data augmentation.}
	\vspace{-0.07in}
	\label{tab:perception_downstream}
	\resizebox{\columnwidth}{!}{
		\begin{tabular}{lccccc}
			\toprule
			\multirow{1}{*}{\textbf{Methods}} & mAP$\uparrow$ & AP@1m$\uparrow$ & AP@2m$\uparrow$ & AP@4m$\uparrow$ \\
			\midrule
			Real & {0.256} & {0.085} & {0.247} & {0.437}\\
			Real + Sim (3DGS) & 0.258 & {0.097} & 0.246 & 0.430\\
			Real + Sim (ours) & \textbf{0.277} & \textbf{0.105} & \textbf{0.272} & \textbf{0.453} \\
			\bottomrule
		\end{tabular}
	}
	\vspace{-0.1in}
\end{table}

Finally, to study whether simulation data improves 3D detection training, we augment PandaSet training set with
renders generated by 3DGS and \namefull{} at novel views: lateral shifts of 3\,m while maintaining scene content unchanged. We retrain {BEVformer-tiny}~\cite{bevformer} on the union of real and simulated images and evaluate on real held-out data. As reported in Table~9, \name{}-based augmentation yields clear improvements in average precision, whereas augmentation with vanilla 3DGS renders provides no noticeable gain. These findings indicate that high-fidelity, extrapolation-stable simulation is important for effective data augmentation in perception.

\section{Limitations}
\label{sec:limitation}

\name{} has several limitations. First, although \name{} produces robust 3DGS representations, it still exhibits noticeable artifacts under extreme extrapolations (\textit{e.g.}, bird-eye viewpoints in Fig.~\ref{fig:pipeline}).
Second, \name{} does not achieve 360° shape completion for background or objects~\cite{yang2025genassets}: geometry and appearance in heavily occluded or unobserved regions remain under-constrained.
Addressing these challenges is an important direction towards fully unconstrained view synthesis and complete scene reconstruction.

\section{Conclusion}
\label{sec:conclusion}

We introduce \name, a diffusion-guided generalizable enhancer for urban scene reconstruction.
\name takes as input any pre-trained 3D Gaussian representation and fixes the deficiencies within 2 minutes in a generalizable manner.
At the heart of \name are two modules: a one-step diffusion neural fixer that fixes degraded rendered images and a generalizable enhancer that predicts per-Gaussian
residuals
to enhance the representation at novel views.
Additionally, we also show that by adapting the enhancer for scene reconstruction from scratch,
we obtain a generalizable reconstruction model that can robustly
reconstruct
the scene within 60s.
Experiments show that \name outperforms existing methods in both quality and efficiency and benefits various downstream tasks, enabling robust and scalable sensor simulation for autonomous driving.

\bibliographystyle{plain}
\bibliography{main.bib}

@String(CVPR= {IEEE Conf. Comput. Vis. Pattern Recog.})

@String(ECCV= {Eur. Conf. Comput. Vis.})

@String(TOG= {ACM Trans. Graph.})

@String(ICLR = {Int. Conf. Learn. Represent.})

@String(CVPR  = {CVPR})

@String(ECCV  = {ECCV})

@String(TOG   = {ACM TOG})

@String(ICLR  = {ICLR})

@inproceedings{3dgs,
	author = {Kerbl, Bernhard and Kopanas, Georgios and Leimk{\"u}hler, Thomas and Drettakis, George},
	booktitle = {TOG},
	title = {{3D} Gaussian Splatting for Real-Time Radiance Field Rendering},
	year = {2023}
}

@inproceedings{nerf,
    title={NeRF: Representing Scenes as Neural Radiance Fields for View Synthesis},
    author={Ben Mildenhall and Pratul P. Srinivasan and Matthew Tancik and Jonathan T. Barron and Ravi Ramamoorthi and Ren Ng},
    year={2020},
    booktitle={ECCV},
}

@inproceedings{carla,
    author    = {Alexey Dosovitskiy and
                Germ{\'{a}}n Ros and
                Felipe Codevilla and
                Antonio M. L{\'{o}}pez and
                Vladlen Koltun},
    title     = {{CARLA:} An Open Urban Driving Simulator},
    booktitle = {CoRL},
    year      = {2017},
}

@inproceedings{airsim,
    title        = {Airsim: High-fidelity visual and physical simulation for autonomous vehicles},
    author       = {Shah, Shital and Dey, Debadeepta and Lovett, Chris and Kapoor, Ashish},
    booktitle    = {Field and service robotics},
    year         = {2018},
}

@inproceedings{yang2023unisim,
    title={UniSim: A Neural Closed-Loop Sensor Simulator},
    author={Yang, Ze and Chen, Yun and Wang, Jingkang and Manivasagam, Sivabalan and Ma, Wei-Chiu and Yang, Anqi Joyce and Urtasun, Raquel},
    booktitle={CVPR},
    year={2023}
}

@inproceedings{tonderski2024neurad,
    author = {Tonderski, Adam and Lindstr{\"o}m, Carl and Hess, Georg and Ljungbergh, William and Svensson, Lennart and Petersson, Christoffer},
    booktitle = {CVPR},
    title = {{NeuRAD}: Neural rendering for autonomous driving},
    year = {2024}
}

@inproceedings{yan2024street,
    title={Street Gaussians: Modeling Dynamic Urban Scenes with Gaussian Splatting},
    author={Yunzhi Yan and Haotong Lin and Chenxu Zhou and Weijie Wang and Haiyang Sun and Kun Zhan and Xianpeng Lang and Xiaowei Zhou and Sida Peng},
    booktitle={ECCV},
    year={2024}
}

@inproceedings{lidarsim,
    title={Lidarsim: Realistic lidar simulation by leveraging the real world},
    author={Manivasagam, Sivabalan and Wang, Shenlong and Wong, Kelvin and Zeng, Wenyuan and Sazanovich, Mikita and Tan, Shuhan and Yang, Bin and Ma, Wei-Chiu and Urtasun, Raquel},
    booktitle={CVPR},
    year={2020}
}

@inproceedings{wang2022cadsim,
    title={CADSim: Robust and Scalable in-the-wild 3D Reconstruction for Controllable Sensor Simulation},
    author={Wang, Jingkang and Manivasagam, Sivabalan and Chen, Yun and Yang, Ze and B{\^a}rsan, Ioan Andrei and Yang, Anqi Joyce and Ma, Wei-Chiu and Urtasun, Raquel},
    booktitle={CoRL},
    year={2022}
}

@inproceedings{chen2025g3r,
    author = {Chen, Yun and Wang, Jingkang and Yang, Ze and Manivasagam, Sivabalan and Urtasun, Raquel},
    booktitle = {ECCV},
    title = {G3R: Gradient Guided Generalizable Reconstruction},
    year = {2025}
}

@article{chen2023periodic,
	title={Periodic vibration gaussian: Dynamic urban scene reconstruction and real-time rendering},
	author={Chen, Yurui and Gu, Chun and Jiang, Junzhe and Zhu, Xiatian and Zhang, Li},
	journal={arXiv},
	year={2023}
}

@inproceedings{chen2024omnire,
    author = {Chen, Ziyu and Yang, Jiawei and Huang, Jiahui and de Lutio, Riccardo and Esturo, Janick Martinez and Ivanovic, Boris and Litany, Or and Gojcic, Zan and Fidler, Sanja and Pavone, Marco and others},
    booktitle = {ICLR},
    title = {OmniRe: Omni Urban Scene Reconstruction},
    year = {2024}
}

@article{onestepdiffusion,
    title={One-Step Image Translation with Text-to-Image Models},
    author={Gaurav Parmar and Taesung Park and Srinivasa Narasimhan and Jun-Yan Zhu},
    year={2024},
    journal = {arXiv},
}

@inproceedings{wu2025difix3d,
    title={Difix3D+: Improving 3D Reconstructions with Single-Step Diffusion Models},
    author={Jay Zhangjie Wu and Yuxuan Zhang and Haithem Turki and Xuanchi Ren and Jun Gao and Mike Zheng Shou and Sanja Fidler and Zan Gojcic and Huan Ling},
    booktitle={CVPR},
    year={2025}
}

@inproceedings{freevs,
    title={Freevs: Generative view synthesis on free driving trajectory},
    author={Wang, Qitai and Fan, Lue and Wang, Yuqi and Chen, Yuntao and Zhang, Zhaoxiang},
    booktitle={ICLR},
    year={2025}
}

@inproceedings{dreamdrive,
    author    = {Mao, Jiageng and Li, Boyi and Ivanovic, Boris and Chen, Yuxiao and Wang, Yan and You, Yurong and Xiao, Chaowei and Xu, Danfei and Pavone, Marco and Wang, Yue},
    title     = {DreamDrive: Generative 4D Scene Modeling from Street View Images},
    booktitle   = {ICRA},
    year      = {2025},
}

@inproceedings{vegs,
    title={VEGS: View Extrapolation of Urban Scenes in 3D Gaussian Splatting using Learned Priors},
    author={Sungwon Hwang and Min-Jung Kim and Taewoong Kang and Jayeon Kang and Jaegul Choo},
    year={2024},
    booktitle={ECCV},
}

@inproceedings{scube,
    title={SCube: Instant Large-Scale Scene Reconstruction using VoxSplats},
    author={Ren, Xuanchi and Lu, Yifan and Liang, Hanxue and Wu, Jay Zhangjie and Ling, Huan and Chen, Mike and Fidler, Sanja annd Williams, Francis and Huang, Jiahui},
    booktitle={NeurIPS},
    year={2024},
}

@inproceedings{streetcrafter,
    title={StreetCrafter: Street View Synthesis with Controllable Video Diffusion Models},
    author={Yan, Yunzhi and Xu, Zhen and Lin, Haotong and Jin, Haian and Guo, Haoyu and Wang, Yida and Zhan, Kun and Lang, Xianpeng and Bao, Hujun and Zhou, Xiaowei and Peng, Sida},
    booktitle={CVPR},
    year={2025},
}

@inproceedings{freesim,
    title={FreeSim: Toward Free-viewpoint Camera Simulation in Driving Scenes},
    author={Lue Fan and Hao Zhang and Qitai Wang and Hongsheng Li and Zhaoxiang Zhang},
    year={2025},
    booktitle={CVPR},
}

@article{mudg,
    title={MuDG: Taming Multi-modal Diffusion with Gaussian Splatting for Urban Scene Reconstruction},
    author={Zou, Yingshuang and Ding, Yikang and Zhang, Chuanrui and Guo, Jiazhe and Li, Bohan and Lyu, Xiaoyang and Tan, Feiyang and Qi, Xiaojuan and Wang, Haoqian},
    journal={arXiv},
    year={2025}
}

@article{recondreamer,
    title={ReconDreamer: Crafting World Models for Driving Scene Reconstruction via Online Restoration},
    author={Chaojun Ni and Guosheng Zhao and Xiaofeng Wang and Zheng Zhu and Wenkang Qin and Guan Huang and Chen Liu and Yuyin Chen and Yida Wang and Xueyang Zhang and Yifei Zhan and Kun Zhan and Peng Jia and Xianpeng Lang and Xingang Wang and Wenjun Mei},
    journal={arxiv},
    year={2024},
}

@article{recondreamer++,
    title={ReconDreamer++: Harmonizing Generative and Reconstructive Models for Driving Scene Representation},
    author={Guosheng Zhao and Xiaofeng Wang and Chaojun Ni and Zheng Zhu and Wenkang Qin and Guan Huang and Xingang Wang},
    journal={arxiv},
    year={2025},
}

@article{dist4d,
    title={DiST-4D: Disentangled Spatiotemporal Diffusion with Metric Depth for 4D Driving Scene Generation},
    author={Jiazhe Guo and Yikang Ding and Xiwu Chen and Shuo Chen and Bohan Li and Yingshuang Zou and Xiaoyang Lyu and Feiyang Tan and Xiaojuan Qi and Zhiheng Li and Hao Zhao},
    journal={arXiv},
    year={2025}
}

@inproceedings{splatformer,
    title = {SplatFormer: Point Transformer for Robust 3D Gaussian Splatting},
    author = {Chen, Yutong and Mihajlovic, Marko and Chen, Xiyi and Wang, Yiming and Prokudin, Sergey and Tang, Siyu},
    booktitle = {ICLR},
    year = {2025}
}

@article{objaverseXL,
    title={Objaverse-XL: A Universe of 10M+ 3D Objects},
    author={Matt Deitke and Ruoshi Liu and Matthew Wallingford and Huong Ngo and
            Oscar Michel and Aditya Kusupati and Alan Fan and Christian Laforte and
            Vikram Voleti and Samir Yitzhak Gadre and Eli VanderBilt and
            Aniruddha Kembhavi and Carl Vondrick and Georgia Gkioxari and
            Kiana Ehsani and Ludwig Schmidt and Ali Farhadi},
    journal={arXiv},
    year={2023}
}

@inproceedings{wonder3d,
    title={Wonder3D: Single Image to 3D using Cross-Domain Diffusion},
    author={Long, Xiaoxiao and Guo, Yuan-Chen and Lin, Cheng and Liu, Yuan and Dou, Zhiyang and Liu, Lingjie and Ma, Yuexin and Zhang, Song-Hai and Habermann, Marc and Theobalt, Christian and others},
    booktitle={CVPR},
    year={2024}
}

@article{sdturbo,
    title={Adversarial Diffusion Distillation},
    author={Axel Sauer and Dominik Lorenz and Andreas Blattmann and Robin Rombach},
    year={2023},
    journal={arXiv},
}

@inproceedings{lora,
    title={LoRA: Low-Rank Adaptation of Large Language Models},
    author={Edward J Hu and yelong shen and Phillip Wallis and Zeyuan Allen-Zhu and Yuanzhi Li and Shean Wang and Lu Wang and Weizhu Chen},
    booktitle={ICLR},
    year={2022},
}

@inproceedings{pandaset,
    title={Pandaset: Advanced sensor suite dataset for autonomous driving},
    author={Xiao, Pengchuan and Shao, Zhenlei and Hao, Steven and Zhang, Zishuo and Chai, Xiaolin and Jiao, Judy and Li, Zesong and Wu, Jian and Sun, Kai and Jiang, Kun and others},
    booktitle={ITSC},
    year={2021},
}

@inproceedings{splatad,
    title={SplatAD: Real-Time Lidar and Camera Rendering with 3D Gaussian Splatting for Autonomous Driving},
    author={Hess, Georg and Lindstr{\"o}m, Carl and Fatemi, Maryam and Petersson, Christoffer and Svensson, Lennart},
    booktitle={CVPR},
    year={2025}
}

@inproceedings{fid,
    title={On Aliased Resizing and Surprising Subtleties in GAN Evaluation},
    author={Parmar, Gaurav and Zhang, Richard and Zhu, Jun-Yan},
    booktitle={CVPR},
    year={2022}
}

@inproceedings{peng2025desire,
	title={Desire-gs: 4d street gaussians for static-dynamic decomposition and surface reconstruction for urban driving scenes},
	author={Peng, Chensheng and Zhang, Chengwei and Wang, Yixiao and Xu, Chenfeng and Xie, Yichen and Zheng, Wenzhao and Keutzer, Kurt and Tomizuka, Masayoshi and Zhan, Wei},
	booktitle={CVPR},
	year={2025}
}

@inproceedings{dreamfusion,
	title={DreamFusion: Text-to-3D using 2D Diffusion},
	author={Ben Poole and Ajay Jain and Jonathan T. Barron and Ben Mildenhall},
	booktitle={ICLR},
	year={2023},
}

@inproceedings{reconfusion,
	title={Reconfusion: 3d reconstruction with diffusion priors},
	author={Wu, Rundi and Mildenhall, Ben and Henzler, Philipp and Park, Keunhong and Gao, Ruiqi and Watson, Daniel and Srinivasan, Pratul P and Verbin, Dor and Barron, Jonathan T and Poole, Ben and others},
	booktitle={CVPR},
	year={2024}
}

@misc{wu2019detectron2,
    author =       {Yuxin Wu and Alexander Kirillov and Francisco Massa and
                    Wan-Yen Lo and Ross Girshick},
    title =        {Detectron2},
    howpublished = {\url{https://github.com/facebookresearch/detectron2}},
    year =         {2019}
}

@article{bevformer,
    title={BEVFormer v2: Adapting Modern Image Backbones to Bird's-Eye-View Recognition via Perspective Supervision},
    author={Chenyu Yang and Yuntao Chen and Haofei Tian and Chenxin Tao and Xizhou Zhu and Zhaoxiang Zhang and Gao Huang and Hongyang Li and Y. Qiao and Lewei Lu and Jie Zhou and Jifeng Dai},
    journal={ArXiv},
    year={2022},
}

@inproceedings{yang2025genassets,
    title     = {GenAssets: Generating in-the-wild 3D Assets in Latent Space},
    author    = {Ze Yang and Jingkang Wang and Haowei Zhang and Sivabalan Manivasagam and Yun Chen and Raquel Urtasun},
    booktitle = {CVPR},
    year      = {2025},
}

@inproceedings{wang2025flux4d,
	title={Flux4D: Flow-based Unsupervised 4D Reconstruction},
	author={Jingkang Wang and Henry Che and Yun Chen and Ze Yang and Lily Goli and Sivabalan Manivasagam and Raquel Urtasun},
	booktitle={NeurIPS},
	year={2025},
}

@inproceedings{wang2021advsim,
	title={Advsim: Generating safety-critical scenarios for self-driving vehicles},
	author={Wang, Jingkang and Pun, Ava and Tu, James and Manivasagam, Sivabalan and Sadat, Abbas and Casas, Sergio and Ren, Mengye and Urtasun, Raquel},
	booktitle={CVPR},
	year={2021}
}

\addtolength{\textheight}{-12cm}

\end{document}